\documentclass[letterpaper, 10 pt, journal, twoside]{IEEEtran}
\usepackage[T1]{fontenc}
\usepackage{newtxtext}
\usepackage{amsmath,amssymb}
\usepackage{amsthm}
\usepackage{algorithmic}
\usepackage{graphicx}
\usepackage{subfigure}
\usepackage{textcomp}
\usepackage{xcolor}
\usepackage{bm}
\usepackage{url}
\newtheorem{theorem}{Theorem}
\usepackage{float}

\ifCLASSINFOpdf
\else
\fi

\author{Dimitria Silveria$^{1}$, Kleber Cabral$^{2}$, Peter T. Jardine$^2$, and Sidney Givigi$^{2}$% <-this % stops a space
\thanks{Manuscript received: July, 05, 2025; Revised September, 01, 2025; Accepted November, 05, 2025.}
\thanks{This paper was recommended for publication by Editor Giuseppe Loianno Name evaluation of the Associate Editor and Reviewers' comments.} %Use only for final RAL version
\thanks{$^{1}$D. Silveria is with the Department of Electrical and Computer Engineering, and the Ingenuity Labs Research Institute 
        Queen's University, 19 Union St, Kingston, ON K7L 3N9
        {\tt\small dimitria.s@queensu.ca}}%
\thanks{$^{2}$  K.\ Cabral, P. Jardine, and S.\ Givigi are with the School of Computing and the Ingenuity Labs Research Institute, 
        Queen's University, Kingston, ON K7L 3N6 Canada 
        {\tt\small kleber.cabral@queensu.ca, p.jardine@queensu.ca, sidney.givigi@queensu.ca}}%
\thanks{Digital Object Identifier (DOI):see top of this page}
}

\title{Decentralized Swarm Control via SO(3) Embeddings for 3D Trajectories}
\begin{document}
\maketitle
\markboth{IEEE Robotics and Automation Letters. Preprint Version. November, 2025}
{Silveria \MakeLowercase{\textit{et al.}}: Decentralized Swarm Control via SO(3)
Embeddings for 3D Trajectories}





\begin{abstract}
This paper presents a novel decentralized approach for achieving emergent behavior in multi-agent systems with minimal information sharing. Based on prior work in simple orbits, our method produces a broad class of stable, periodic trajectories by stabilizing the system around a Lie group-based geometric embedding. Employing the Lie group SO(3), we generate a wider range of periodic curves than existing quaternion-based methods. Furthermore, we exploit SO(3) properties to eliminate the need for velocity inputs, allowing agents to receive only position inputs. We also propose a novel phase controller that ensures uniform agent separation, along with a formal stability proof. Validation through simulations and experiments showcases the method's adaptability to complex low-level dynamics and disturbances. 
\end{abstract}

% \begin{IEEEkeywords}
% IEEE, IEEEtran, journal, \LaTeX, paper, template.
% \end{IEEEkeywords}
% \begin{IEEEkeywords}
% List of keywords (from the RA Letters keyword list)
% \end{IEEEkeywords}

% \fi
\IEEEpeerreviewmaketitle

\section{Introduction}
% for IEEE journal papers produced under \LaTeX\ using
% IEEEtran.cls version 1.8b and later.
\IEEEPARstart{A}{ swarm} is a decentralized form of multi-agent system (MAS) that displays emergent behavior \textemdash that is, complex behaviors arising from local interactions governed by simple rules without centralized coordination~\cite{Debie_swarm-survey_2023}. Swarm agents are often robotic platforms such as uncrewed aerial vehicles (UAVs) used in various domains, including entertainment, surveillance, and defense. 

This paper addresses the challenge of generating stable, closed 3D formations around a fixed point for UAVs using only local position information. Such formations are relevant in dynamic capture, surveillance, and mobbing scenarios~\cite{travis}, and relate to applications such as lattice formation~\cite{HU2022110235}, encirclement~\cite{hafez-2015}, epitrochoidal motion~\cite{Fedele_hypotrochoidal_2023}, target enclosing~\cite{Sinha_enclosing_2022}, and other dynamic patterns~\cite{DONG-2020}.

Existing approaches often rely on consensus-based algorithms. For example, \cite{engproc_pso-encirclement_2023} uses consensus control and heading error compensation for 2D circular trajectories, with particle swarm optimization (PSO) applied to tune controller gains. However, this method scales poorly, lacks real-world validation, and is vulnerable to agent loss. Similarly, \cite{Pichierri_distributed-feedback_2024} applies consensus-based optimization for simulated circular patrolling.

% Potential-based methods offer an alternative by guiding agents through 3D circular paths while ensuring collision avoidance~\cite{Carolis_encirclement-agriculture_2023}. Yet, these are often limited to homogeneous agents with simple dynamics and also lack physical testing.

Prior work confines trajectories to circles due to their simplicity and inherent collision avoidance through phase spacing~\cite{Litimein_circular-survey_2021}. Some exceptions exist, such as~\cite{Fedele_hypotrochoidal_2023}, which introduces 3D hypotrochoidal and epitrochoidal trajectories. However, these paths depend on initial conditions, assume double integrator dynamics, lack collision guarantees, and are not validated in physical systems.

To overcome the above limitations, \cite{JARDINE_2025_embedding} introduced an embedding approach where agents follow virtual circular paths with uniform phase spacing enforced via angular velocity control. The circular embedding is a projection of the multi-agent coordination problem into a lower-dimensional space, enabling the use of linear control laws that would otherwise not be directly applicable. Actual trajectories are derived through quaternion-based 3D deformations of this circular trajectory. While this enables the representation of certain complex 3D shapes and adapts to agent loss or addition, it suffers from ambiguities in even-sized swarms: quaternion rotation may map two antipodal agents to the same point, thereby causing a risk of collision. Additionally, this approach is limited to generating lemniscate curves and requires both position and velocity inputs, which constrains the range of applications. Moreover, the previous work restricted its validation to simulation. The authors identify these shortcomings as areas for future work. 

% \subsection{Contributions}

This paper leverages the concept of circular embedding proposed in~\cite{JARDINE_2025_embedding} and makes the following contributions:

\begin{itemize}
    \item A Lie group SO(3)-based trajectory deformation method that generalizes better than quaternion-based approaches, enabling linear control in lower-dimensional space while supporting a greater range of 3D formations, not achieved in the previous literature.
    \item A resolution to the problem of antipodal collisions identified in previous work.
    \item A position-only reference generation technique, allowing for broader applicability to UAVs without velocity control (e.g., Crazyflies\cite{crazyflie}).
    \item A scalable, electrostatic repulsion-inspired phase separation controller that maintains uniform angular separation with formal guarantees of stability.
    \item Experimental validation on physical quadcopters, bridging the gap between theory and practice and demonstrating robustness to unmodeled dynamics.

\end{itemize}

% \subsection{Outline}
The remainder of the paper is organized as follows. Section~\ref{sec:so(3)} introduces the relevant Lie group concepts. Section~\ref{sec:problem_formulation} details the controller design for generating periodic formations. Stability analysis is presented in Section~\ref{sec:stability}. Section~\ref{sec:experiments} outlines the experimental setup for hardware validation, while Section~\ref{sec:results} reports results for both simulation and physical experiments. Finally, Section~\ref{sec:conclusion} concludes the paper and outlines directions for future work.

\section{Preliminaries}\label{sec:so(3)}

This section provides a concise overview of the mathematical foundations relevant to this paper, focusing on the Lie group $\text{SO}(3)$ and its associated Lie algebra $\mathfrak{so}(3)$. For a more comprehensive treatment, see \cite{book,more-lie,martin2021lie}.

The group comprising all 3D rotation matrices, including the identity, is defined as $\text{SO}(3) :=  \{ \bm{R} \in \mathbb{R}^{3 \times 3} \mid \bm{R}^\top \bm{R} = \bm{I} \text{ and } \det(\bm{R}) = +1\}$. Its tangent space at the identity, $T_1$SO(3), forms the Lie algebra, which consists of skew-symmetric matrices and is defined as $\mathfrak{so}(3) := \{ \bm{\Omega} \in \mathbb{R}^{3 \times 3} \mid \bm{\Omega}^\top = -\bm{\Omega} \}$.

Using $\text{SO}(3)$ to represent rotations offers notable advantages over unit quaternions. Quaternion rotations are sign-ambiguous (i.e., $q$ and $-q$ encode the same rotation). In contrast, $\text{SO}(3)$ avoids such ambiguities and is free of numerical drift, eliminating the need for periodic renormalization~\cite{Hall_quantum_2013}.

Rotations in $\mathfrak{so}(3)$ can be expressed via the hat operator $(\cdot)^\wedge$, which maps a 3D angular velocity vector $\boldsymbol{\omega} = [\omega_x, \omega_y, \omega_z]^\top$ into a skew-symmetric matrix: 
\begin{equation}
    \bm{\Omega} = \boldsymbol{\omega}^{\wedge}= \begin{pmatrix} 0 & -\omega_z & \omega_y \\ \omega_z & 0 & -\omega_x \\ -\omega_y & \omega_x & 0 \end{pmatrix}.
\end{equation}
The inverse mapping is given by the vee operator $(\cdot)^{\vee}$.
% Another benefit of $\mathfrak{so}(3)$ lies in its linearity: whereas composition in $\text{SO}(3)$ involves matrix multiplication (e.g., $\bm{R} = \bm{R}_1 \bm{R}_2 \in \text{SO}(3)$), the elements of $\mathfrak{so}(3)$ can be added directly as in $\bm{A} + \bm{B} \in \mathfrak{so}(3)$, simplifying computations for small-angle approximations. Additionally, Lie algebra representations 

The exponential map converts elements of $\mathfrak{so}(3)$ into $\text{SO}(3)$:
\begin{equation}
    \bm{R} = e^{(\bm{\Omega})}.
    \label{eq:exponencial-map}
\end{equation}
% = \bm{I} + \bm{\Omega} + \frac{\bm{\Omega}^2}{2!} + \frac{\bm{\Omega}^3}{3!} + \dots

In this paper, $\mathfrak{so}(3)$ serves as the mathematical foundation for a novel control strategy to generate periodic trajectories in multi-agent robotic swarms.
%This Lie algebra builds on the benefits of $\text{SO}(3)$, providing advantages over quaternion- and rotation matrix-based representations.
Working in $\mathfrak{so}(3)$ allows us to specify unconstrained parametric deformations;
%The first advantage arises from the unconstrained magnitude of elements of $\mathfrak{so}(3)$; 
this contrasts with unit quaternions and rotation matrices, which must satisfy norm constraints. Consequently, the elements in $\bm{\Omega} \in \mathfrak{so}(3)$ may contain any values in $\mathbb{R}$, or be expressed as functions of other system variables (e.g., parametric equations).
If the matrix $\bm{\Omega}$ is mapped to a rotation matrix using \eqref{eq:exponencial-map}, the resulting rotation can be used to \textit{deform} a 2D circular trajectory into a more complex 3D shape, similar to the quaternion-based method in \cite{JARDINE_2025_embedding}. 
%This process is depicted in the next section.
This flexibility in the range of values of $\bm{\Omega}$ expands the range of achievable 3D trajectories compared to quaternion-based approaches. 

% each point in a circle can be uniquely represented by a radius $r$ and an angle $\phi$ (polar coordinates).
% it is easier to define, in $\mathfrak{so}(3)$, the operations that map a point from a circle to the 3D space, as functions of $\phi$. 

% These operations can be converted into rotations using the operation in \eqref{eq:exponencial-map}. Therefore, the representation in $\text{SO}(3)$ increases the number of 3D shapes that can be generated by distorting a circle, compared to the quaternion representation proposed in \cite{JARDINE_2025_embedding}.

The swarm pipeline proposed in this paper controls the motion of agents in 3D by regulating their angular velocity in a 2D circular plane. The angular velocity is represented as a skew-symmetric matrix $\mathfrak{so}(3)$ and mapped to a rotation matrix using \eqref{eq:exponencial-map}. 
Another advantage of this formulation is that it simplifies the low-level controller, requiring only position references rather than position and velocity as in \cite{JARDINE_2025_embedding}. This process is discussed in detail in Section~\ref{sec:position-control}.

Additionally, the position noise and uncertainties can be encoded in the Lie algebra, where rotations are defined as linear disturbances that are projected back to the group space using the exponential map. This approach of encoding uncertainties in Lie algebras is widely used in state estimation, particularly with the invariant Kalman filter \cite{Xu_invariant_2023}.

% This algebraic structure offers several computational benefits. Unlike the nine parameters required for a full rotation matrix, $\mathfrak{so}(3)$ uses only three, enabling a more compact and intuitive rotation representation. Moreover, the matrix $\bm{\Omega}$ encodes both rotation angle and axis~\cite{martin2021lie}, avoiding gimbal lock issues commonly associated with Euler angles~\cite{murray2017mathematical}.
\section{Controller Design}\label{sec:problem_formulation}

This section outlines the control system architecture for a swarm using $\text{SO}(3)$ rotations, where the Lie Group embedding acts as a form of feedback linearization, enabling standard control of the desired periodic motion.

Each agent $i$ is modeled as a particle in free space, characterized by its position ($\bm{x}_i~\in~\mathbb{R}^3$), velocity ($\bm{v}_i~\in~\mathbb{R}^3$), and control input ($\bm{u}_i~\in~\mathbb{R}^3$). The agent's dynamics is governed by the following second-order system:
\begin{equation} \label{eq:dyn}
\begin{matrix}
\dot{\bm{x}}_i = \bm{v}_i, \\
\dot{\bm{v}}_i = \bm{u}_i,
\end{matrix}
\end{equation}
\noindent where $\bm{u}_i=\ddot{\bm{x}}_i$ is the agent's acceleration input. 

\subsection{Lie Group Embedding}\label{sec:embedding}

Let us introduce a geometric embedding based on Lie group theory. Consider a circular trajectory of radius $r$ and phase angle $\phi\in[0,2\pi)$. Let $\omega_x(\phi)$ and $\omega_y(\phi)$ define parametric angular velocities that deform this trajectory over time about the $X$ and $Y$ axes, respectively. These deformations are encoded in the following angular velocities vector in $\mathfrak{so}(3)$:% that represents this rotation is given by:
\begin{equation}
    \bm{\omega}_{xy}(\phi) = 
    \begin{bmatrix}
       \omega_x(\phi) & \omega_y(\phi) & 0
    \end{bmatrix}^{\top},
    \qquad
    \bm{\Omega}_{xy}(\phi) = 
    \boldsymbol{\omega}_{xy}(\phi)^\wedge.
    \label{eq:skew-x-y}
\end{equation}

The exponential map transforms this Lie algebra element into a rotation matrix in $\text{SO}(3)$:
\begin{equation}
     \bm{R}
    %= e^{\bm{\omega}_{xy}^\wedge}.
    = e^{\bm{\Omega}_{xy}(\phi)}.
    \label{eq:exponential}
\end{equation}
\noindent where $\bm{R}$ $\in$ $SO(3)$ is the time-varying rotation that deforms the base circular trajectory into a 3D curve. Applying this transformation yields the deformed trajectory:
 \begin{equation} \label{eq:rotation}
{\bm{x}} = \bm{R}\hat{\bm{x}},
\end{equation}
where 
\begin{equation}
    \hat{\bm{x}} = 
    \begin{bmatrix}
        r\cos{(\phi)} && r\sin{(\phi)} && 0
    \end{bmatrix}^\top
    \label{eq:embedding-coords}
\end{equation}

\noindent is the original circular trajectory in the $XY$-plane.

\begin{figure}%[!ht]
    \centering
    % First row
    \subfigure[1][$\omega_x= s \sin(6\phi)\cos(6\phi)$, $\omega_y= s$, $\dot\phi_d=0.8$ rad/s, $s=0.3$]{%
        \includegraphics[width=0.45\linewidth]{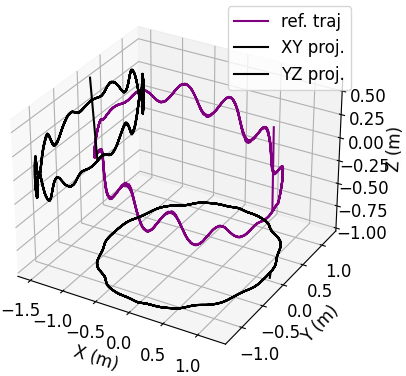}%
        \label{sub-fig:3D-diff-shapes-a}
    }
    \hfill
    \subfigure[2][$\omega_x= s \sin(\phi)\cos(\phi)$, $\omega_y=0$, $\dot\phi_d=2$ rad/s, $s=1$]{%
        \includegraphics[width=0.45\linewidth]{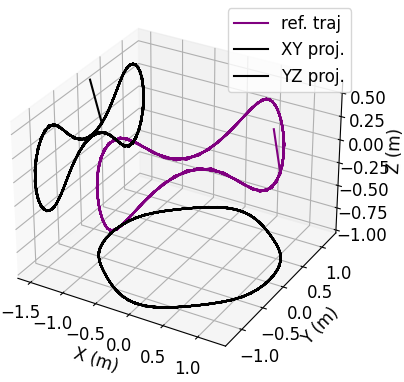}%
        \label{sub-fig:3D-diff-shapes-b}
    }
    \vskip\baselineskip
    % Second row
    \subfigure[3][$\omega_x= s\cos(2\phi)$, $\omega_y= s\cos^2(\phi)$, $\dot\phi_d=0.5$ rad/s, $s=0.6$]{%
        \includegraphics[width=0.45\linewidth]{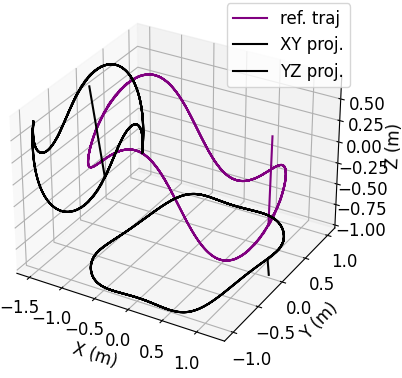}%
        \label{sub-fig:3D-diff-shapes-c}
    }
    \hfill
    \subfigure[4][$\omega_x= s\cos(3\phi)\sin(\phi)$, $\omega_y=0.5s$, $\dot\phi_d=1.8$ rad/s, $s=0.9$]{%
        \includegraphics[width=0.45\linewidth]{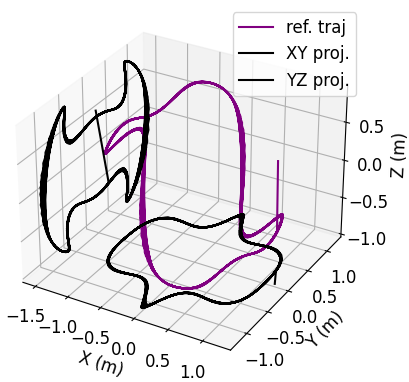}%
        \label{sub-fig:3D-diff-shapes-d}
    }
    \caption{Examples of 3D geometries generated using the method described in Section~\ref{sec:embedding} along with $XY$ and $YZ$ projections.}
    \label{fig:3D-diff-shapes}
\end{figure}

Fig.~\ref{fig:3D-diff-shapes} shows four distinct trajectory deformations using this method. For example, the distortion shown in Fig.~\ref{sub-fig:3D-diff-shapes-b} results in a Dumbbell-shaped projection in the $YZ$ plane -- analogous to the $XY$-plane Dumbbell structure in~\cite{JARDINE_2025_embedding} -- demonstrating the versatility of the approach.

The distortion factor $s$ governs the degree of deformation: $s=0$ corresponds to an undistorted circle (i.e., a real-world circular trajectory), while larger values of $s$ induce increasingly complex 3D structures. Parameters for each shape are provided in the figure captions. 
\subsection{Controller Overview}\label{sec:controller}

Consider an agent $i$ following a 3D trajectory derived via the rotation~\eqref{eq:rotation} from a circular reference path %(i.e., the embedding) 
with radius $r_d$ and phase $\phi_i = (\omega_{z,d,i})t$, rotating about the $Z$ axis. The virtual agent $\hat{\bm{x}}_i$ is defined by mapping the agent's position into the circular embedding via the inverse of \eqref{eq:rotation}:
\begin{equation} 
  \hat{\bm{x}}_i = \bm{R}_i{^\top}\bm{x}_i.
  \label{eq:embedding-induction}
\end{equation}

To achieve the desired behavior, we regulate $\hat{\bm{x}}_i$ to asymptotically track the reference circular trajectory: 
 \begin{equation} \label{eq:equival}
	\lim_{t\to\infty} \hat{\bm{x}}_i(t)  
	=
	\begin{bmatrix}
	r_d\cos(\omega_{z,d}t) &
	r_d\sin(\omega_{z,d}t) &
	0
	\end{bmatrix}^\top,
\end{equation}
\noindent where $\omega_{z,d}$ is the common desired angular velocity for all agents, which is constant over time. Regulating this embedding indirectly regulates real-world coordinates $\bm{x}_i$.  This control strategy is implemented in two stages: a phase controller and a position controller. These stages ensure, respectively, the convergence of both angular velocity and radial distance, and are designed to satisfy:
\begin{equation} \label{eq:rtoInf} 
\begin{matrix}
        \lim_{t\to\infty} \omega_{z,d,i}(t) = \omega_{z,d},  & %~\forall~i~\in~N_i^\phi\]
        \lim_{t\to\infty} r_{i}(t) = r_{d}.      
\end{matrix}
\end{equation}
\noindent where $\omega_{z,d,i}(t)$ is the $z$ angular velocity of the virtual agent $i$ at the time instant $t$, and $r_i(t)$ is the radius of this same agent at the same time instant, calculated from the center of the circular embedding. 

The phase controller governs the emergent behavior by adjusting each agent's angular velocity based on the phase differences with its leading and lagging neighbors. This promotes uniform angular spacing and avoids collisions in the embedding. Meanwhile, the linear position controller ensures each agent converges to the appropriate point on the desired trajectory.

\subsection{Phase Control}\label{sec:phase-control}

The phase of agent $i$ is $\phi_i=\arctan2(\hat{x}_{y,i},\hat{x}_{x,i})$, where $\hat{x}_{x,i}$ and $\hat{x}_{y,i}$ are the $x$ and $y$ coordinates of the virtual agent in the circular embedding, according to \eqref{eq:embedding-coords}. Let us define the phase of the leading and lagging agents of $i$ as $\phi_k$, and $\phi_j$, which are obtained in the same way as $\phi_i$, using the respective coordinates of the agents. %By definition $0<\phi_i<2\pi$. 
The phase control law driving the agent's desired angular velocity $\omega_{z,d,i}$ is given by:
\begin{equation}
    \omega_{z,d,i} = \omega_{z,d} + k_{\phi} \left(\frac{1}{\phi_{ki}} + \frac{1}{\phi_{ji}}\right),
    \label{eq:phase-control}
\end{equation}
where $\omega_{z,d}$ is the nominal angular velocity shared by the swarm, $\phi_{ki}=\phi_{i} - \phi_{k}$ is the phase difference between the agent $i$ and its leading agent $k$, $\phi_{ji} = \phi_{i} - \phi_{j}$ is the phase difference between agent $i$ and its lagging agent $j$, and $k_{\phi} > 0$ is a control gain. To ensure consistent angular relationships, both $\phi_{ki}$ and $\phi_{ji}$ are constrained to $(-\pi,\pi)$. 

This control law induces a repulsive behavior between neighboring agents, analogous to electrostatic repulsion between like charges: the repulsion strength increases as the agents approach each other in phase, due to the inverse dependence on $\phi_{ki}$ and $\phi_{ji}$. Notably, $\phi_{ki}$ is always negative (as $k$ leads $i$), while $\phi_{ji}$ is always positive (as $j$ trails $i$). When agent $i$ moves too close to its leader (i.e., $|\phi_{ki}|<|\phi_{ji}|$), the term $1/\phi_{ki}$ dominates, reducing $\omega_{z,d,i}$ and thus slowing the agent down. This causes $\phi_{ki}$ to increase and $\phi_{ji}$ to decrease, restoring balance. Conversely, when $i$ is closer to its follower than its leader, the term $1/\phi_{ji}$ prevails, increasing $\omega_{z,d,i}$ and accelerating the agent to reestablish even spacing.

\subsection{Position Control}\label{sec:position-control}

While the phase controller ensures uniform angular separation, a separate controller is required to stabilize each agent's radial position. This is accomplished by first defining the current desired position %of the virtual agent 
in the circular embedding:
\begin{equation}\label{eqn:xd}
    \hat{\bm{x}}_{d,i}
    =
    \begin{bmatrix}
        r_{d}\cos(\phi_i)  &
        r_{d}\sin(\phi_i) &
        0
    \end{bmatrix}^\top.
\end{equation}
\noindent where $r_d$ is the desired radius and $\phi_i$ is the current phase of agent $i$.
Next, an incremental rotation about the $Z$-axis is constructed using the agent's desired angular velocity $\omega_{z,d,i}$, calculated with \eqref{eq:phase-control}. The corresponding Lie algebra vector is:
\begin{equation}
    \bm{\omega}_{z,i} =
    \begin{bmatrix}
        0 & 0 & \omega_{z,d,i} 
    \end{bmatrix}^\top.
\end{equation}
Applying the exponential map yields the incremental rotation matrix:
\begin{equation}\label{eq:rot_z}
        \Delta \bm{R}_z 
    = e^{\bm{\omega}_{z,i}^{\wedge} dt}.
\end{equation}
\noindent which advances the agent's phase. This rotation is applied to the current desired position $\hat{\bm{x}}_{d,i}$ to form the next desired position $\hat{\bm{x}}^{\prime}_{d,i}$, following
\begin{equation}
    \hat{\bm{x}}^{\prime}_{d,i} = \Delta \bm{R}_z\hat{\bm{x}}_{d,i}.
\end{equation}
\noindent The updated phase is then extracted from this position as $\phi^{\prime}_i=\arctan2(\hat{x}^{\prime}_{y,i},\hat{x}^{\prime}_{x,i})$. The new phase $\phi^{\prime}_i$ is used to obtain the angular velocity matrix $\bm{\Omega}_{xy,i}$, according to \eqref{eq:skew-x-y}. Substituting $\bm{\Omega}_{xy,i}$ into \eqref{eq:exponential} obtains the distortion matrix $\bm{R}_i$.

The desired updated position %in 3D space 
is obtained by applying the embedding transformation $\bm{R}_i$ in~\eqref{eq:exponential}:
\begin{equation} \label{eq:rot}
\bm{x}_{d,i} = \bm{R}_{i}\hat{\bm{x}}^{\prime}_{d,i},
\end{equation}
\noindent where, $\bm{x}_{d,i}$ is given with respect to the center of the embedding.
% where $\bm{R}_{i}$ is the matrix defined in \eqref{eq:exponential} for the agent $i$.

To drive the agent toward this target, we employ a linear proportional-derivative (PD) controller:
% \begin{equation} 
%     \bm{u}_i = k_x\bm{e}_{x} + k_i\int_0^t\bm{e}_{x}d\tau + k_v\frac{d\bm{e}_{x}}{dt} ,
%     \label{eq:controller}
% \end{equation}
\begin{equation} 
    \bm{u}_i = k_x\bm{e}_{x} + k_v\frac{d\bm{e}_{x}}{dt} ,
    \label{eq:controller}
\end{equation}
\noindent where $\bm{e}_{x} = \bm{x}_{d,i} - \bm{x}_{i}$, and $k_x$ and $k_v$ are controller gains selected to ensure stability and desired dynamic performance. 

Although we adopted a PD controller under double integrator dynamics for simplicity, the proposed trajectory generation framework is agnostic to the low-level controller. As demonstrated in Section~\ref{sec:real-results}, it remains effective with more complex systems, including quadcopter platforms.

\section{Stability Analysis}\label{sec:stability}

The stability analysis is divided into two parts: (1) stability of the embedding, and (2) stability of the transformed trajectories.

\subsection{Stability of the Embedding}
The following theorem establishes the Lyapunov stability of the embedding under the phase control law in \eqref{eq:phase-control}.
\begin{theorem}\label{thm:embedding}
    Let $n\geq3$ be the number of agents in a swarm that rotate in a common plane around a fixed point. If each agent's angular velocity $\omega_{z,d,i}$ is governed by the control law in~\eqref{eq:phase-control}, then the system is Lyapunov stable, and all agents converge to a steady-state angular velocity $\omega_{z,d,i} = \omega_{z,d}$, and uniform phase separation (\textit{i.e.}, $360^\circ/n$).
    
\end{theorem}

\begin{proof}
Define the angular phase errors for agent $i$ with respect to its leading ($k$) and lagging ($j$) agents as:
\begin{equation}
\begin{aligned}
    e_{ji} &= \phi_{i} - \phi_{j} = \phi_{ji},\\
    e_{ki} &= \phi_{i} - \phi_{k} = \phi_{ki}.
\end{aligned}
\end{equation}

Consider the following Lyapunov function:
\begin{equation}
    V(e_{ji},e_{ki}) = \sum_{i=1}^n\frac{1}{2}\left(\frac{1}{e_{ji}}+\frac{1}{e_{ki}}\right)^2.
    \label{eq:lyapunov}
\end{equation}
Its time derivative is given by:
\begin{equation}
    \dot{V}(e_{ji},e_{ki}) = \sum_{i=1}^n\left[\left(\frac{1}{e_{ji}}+\frac{1}{e_{ki}}\right)\left(-\frac{\dot{e}_{ji}}{e_{ji}^2}-\frac{\dot{e}_{ki}}{e_{ki}^2}\right)\right].
    \label{eq:lyapunov-derivative}
\end{equation}

At equilibrium, where
$e_{ji} = -e_{ki}$, %$\forall$ $i \leq n$, $i \neq j,k$, 
we observe:

\begin{itemize}
    \item $V(e_{ji},-e_{ji}) = 0$ and $V(e_{ji},e_{ki}) > 0$ elsewhere;
    \item $\dot{V}(e_{ji},-e_{ji}) = 0$ and $\dot{V}(e_{ji},e_{ki}) < 0$ elsewhere
\end{itemize}

To establish that $\dot{V}(e_{ji},e_{ki}) < 0$ outside equilibrium, we analyze the two components of~\eqref{eq:lyapunov-derivative}. From ~\eqref{eq:phase-control}, we derive:% that is, $\left(\frac{1}{e_{ji}}+\frac{1}{e_{ki}}\right)$ and $\left(-\frac{\dot{e}_{ji}}{e_{ji}^2}-\frac{\dot{e}_{ki}}{e_{ki}^2}\right)$ separately.
% From \eqref{eq:phase-control}, we have:
\begin{equation}
     \frac{\omega_{z,d,i} - \omega_{z,d}}{k_{\phi}} =  \left(\frac{1}{e_{ki}} + \frac{1}{e_{ji}}\right).
     \label{eq:phi-diff}
\end{equation}
Taking the time derivative of both sides yields:
\begin{equation}
     \frac{\dot{\omega}_{z,d,i}}{k_{\phi}} =  \left(-\frac{\dot{e}_{ji}}{e_{ji}^2}-\frac{\dot{e}_{ki}}{e_{ki}^2}\right).
     \label{eq:phi-diff-derivative}
\end{equation}
% where $\ddot{\phi}_{d,z,i}$ is the angular acceleration of the agent $i$, and $\dot{e}_{ji}$ and $\dot{e}_{ki}$ are the derivatives of $e_{ji}$ and $e_{ki}$.

Now consider a perturbation where  $|e_{ji}|>|e_{ki}|$, i.e., the agent is closer to its leader than to its follower. Since $e_{ji}>0$ and $e_{ki}<0$, the right-hand side of~\eqref{eq:phi-diff} is negative, causing a decrease in $\omega_{z,d,i}$ (i.e., $\omega_{z,d,i}<\omega_{z,d}$). This slows the agent down, increasing $|e_{ki}|$ and decreasing $|e_{ji}|$, eventually restoring balance (i.e.,   $\omega_{z,d,i}=\omega_{z,d}$). During this process, the angular acceleration $\dot{\omega}_{z,d,i} > 0$, so~\eqref{eq:phi-diff-derivative} is positive, ensuring that $\dot{V}(e_{ji},e_{ki}) < 0$.

A symmetric argument holds for the case $|e_{ji}|<|e_{ki}|$. In all non-equilibrium configurations, the terms in~\eqref{eq:phi-diff} and \eqref{eq:phi-diff-derivative} have opposite signs, guaranteeing that $\dot{V} < 0$.

Therefore, the control law~\eqref{eq:phase-control} ensures Lyapunov stability.
\end{proof}
\subsection{Stability of the Trajectory}\label{sec:embedding-stability}

The next result establishes that the stability of the embedding is preserved when mapped into the 3D trajectory via rotation. This is achieved by showing that the mapping is a one-parameter homeomorphism.
\\
\begin{theorem} \label{thm:stab_homeo}
Let $C:\{\hat{\bm{x}}_1, \hat{\bm{x}}_2, \dots, \hat{\bm{x}}_n~|~ \lVert\hat{\bm{x}}_i\rVert=r~\forall~i,~ \hat{\bm{x}}_i \in \mathbb{R}^{3},~ n>1\}$ represent points uniformly distributed on a circle of radius $r$ in a fixed plane. Define a family of mappings 
\begin{equation*}
    f_i:\hat{\bm{x}}_i\mapsto \bm{x}_i=\bm{R}_i(\phi_i(t))\hat{\bm{x}}_i
\end{equation*}%for each point $\hat{\bm{x}}_i \in C$ using family 
\noindent where $\bm{R}_i(\phi_i(t))\in  SO(3)$ is a time-varying rotation matrix parameterized by the phase $\phi_i(t)\in[0,2\pi)$. If each $\bm{R}_i(\phi_i(t))\in SO(3)$ is continuous, then the family ${f_i}$ defines a one-parameter homeomorphism.
\end{theorem}

\begin{proof}
    A similar result is proven in~\cite{JARDINE_2025_embedding} using quaternion-based rotations. Since both $\text{SO}(3)$ and the unit quaternions $\mathbb{H}^{1}$ are Lie groups with equivalent topological properties, the proof extends naturally to $\text{SO}(3)$.
    
    To confirm that $f_i$ is a homeomorphism, we verify the following properties~\cite{stillwell2008naive}:
    \begin{itemize}
        \item Bijectivity:%$\bm{R}(\phi)$ is bijective:
        
        \begin{itemize}
          % \item $\bm{R}(\phi) \bm{x}_1 = \bm{R}(\phi) \bm{x}_2 \Rightarrow \bm{x}_1 = \bm{x}_2$ 
          \item If $\bm{R}(\phi)\bm{x}_1 = \bm{R}(\phi)\bm{x}_2$, then $\bm{x}_1 = \bm{x}_2$
           % \item  $\forall \bm{x} \in \mathbb{R}^3, \, \exists \hat{\bm{x}} = \bm{R}(\phi)^{-1} \bm{x} \text{ such that } \bm{R}(\phi) \hat{\bm{x}} = \bm{x}$;
           \item For every $\bm{x} \in \mathbb{R}^3$, there exists $\hat{\bm{x}} = \bm{R}^{-1}(\phi)\bm{x}$ such that $\bm{R}(\phi)\hat{\bm{x}} = \bm{x}$.
        \end{itemize}
    \item Continuity of the forward transformation follows from the continuity of matrix multiplication.%Transformations based on $\bm{R}(\phi)$ are continuous because matrix multiplication and transpose are continuous;
    \item Continuity of the inverse transformation also holds due to the continuity of transposition and inversion in $\text{SO}3$.%And any transformation induced by $\bm{R}(\phi)$ has a continuous inverse transformation (again, due to continuity of matrix multiplication and transpose).
    \end{itemize}
    Therefore, the mapping from the embedding to the 3D trajectory is a homeomorphism and preserves stability.
\end{proof}

\section{Experimental Methodology}\label{sec:experiments}
\begin{figure}
    \centering
    \includegraphics[width=0.45\linewidth]{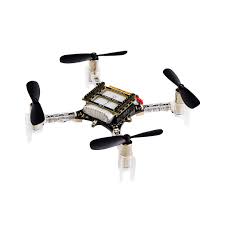}
    \caption{Platform used for physical experiments.}
    \label{fig:crazyflie}
\end{figure}  
\begin{figure*}
    \centering
    \includegraphics[width=\textwidth]{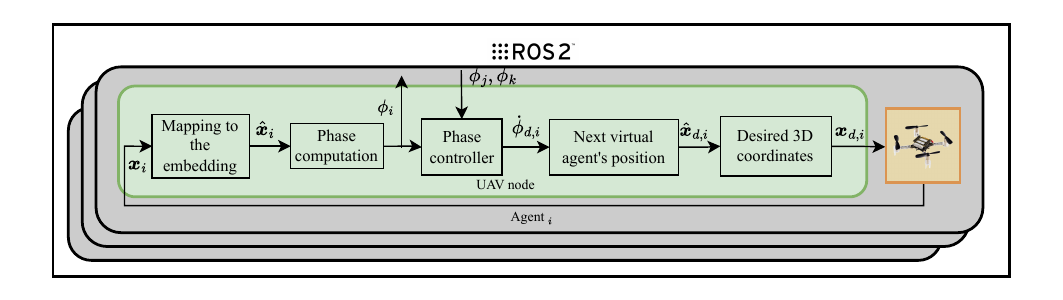}
    \caption{Experimental setup containing the vehicle (orange box), and the control strategy implemented on each UAV (green boxes).}
    \label{fig:experimental setup}
\end{figure*}
The architecture described in Sections~\ref{sec:embedding} and \ref{sec:controller} was implemented in a physical swarm of five UAVs to validate the proposed approach. The experimental platform consisted of Crazyflie quadcopters, by Bitcraze\textsuperscript{\textregistered}, shown in Fig.~\ref{fig:crazyflie}. %\footnote{\url{www.bitcraze.io/products/old-products/crazyflie-2-0/}}

Each UAV was controlled by cascaded PID controllers~\cite{crazyflie_controller} that take position waypoints as input, which justifies the design decision for the trajectory generator to output only position commands, omitting velocity information. The flights were conducted in a $5 \times 7.5 \times 3$ m indoor test area.%, shown in Fig.~\ref{fig:test-area}. %add a figure of the test area%. 

The overall experimental setup is illustrated in Fig. \ref{fig:experimental setup}. The proposed algorithm, represented in the green box, was implemented in Python \footnote{\url{https://github.com/QUARRG/Lie_group_swarm}}, with inter-process communication managed using ROS2. The inertial coordinate frame is fixed at the center of the room at ground level, with the $X-Y$ plane representing the horizontal plane, and the $Z$-axis pointing upward. Position vectors $\bm{x}$ and $\bm{x}_d$ correspond to coordinates in this frame. %\footnote{\url{docs.ros.org/en/humble/index.html}}

As shown in Fig.~\ref{fig:experimental setup}, each agent (gray box) comprises a UAV (orange box) and a ROS2 node (green area) running externally on a ground station, at $10$~Hz. Global position data are provided by a Vicon\textsuperscript{\textregistered} Camera System (not shown in the diagram). This data is fused onboard with internal sensor readings to estimate the UAV's position $\bm{x}_i$, which is then passed to the corresponding ROS2 node. %\footnote{\url{www.vicon.com}}

Each ROS2 node also receives phase values ($\phi_k,\phi_j$)  from its leading and lagging agents, computes the desired 3D target position $\bm{x}_{d,i}$ using the control algorithm, and sends it to the UAV over a radio interface. Communication with the Crazyflie hardware is handled by a dedicated Python library.

Although the control computations were performed on an external computer for implementation convenience, the distributed nature of the proposed architecture permits deployment on embedded systems, such as the Crazyflie's STM32F405 microcontroller or another platform of choice.%\footnote{\url{www.st.com/en/microcontrollers-microprocessors/stm32f405-415.html}}

Each UAV was aware of its leading and lagging agents (e.g., UAV 2 recognized UAV 1 as its leader and UAV 3 as its follower). An external node, with access to the Vicon coordinates, assigned the numbers to each UAV. 

\section{Results}\label{sec:results}

This section provides results from both simulation and physical experiments.

\subsection{Simulation with 50 Agents}\label{sec:sim-results}

The simulation results demonstrate the effectiveness of the proposed pipeline for a 50-agent swarm. The phase controller discussed in Section~\ref{sec:phase-control} ran independently for each agent and was responsible for coordinating the swarm and avoiding collisions between agents. Regardless of the number of agents in the swarm, this local controller required only the phase position of two neighbouring agents. For this reason, communication requirements did not change with an increase in the swarm's size. Therefore --- as long as the trajectory is large enough to fit the agents ---  the approach scales well to larger swarms. For example, let us consider the case where each agent is a sphere with radius $r_a$ such that $r_a$ comprises the dimensions of the agent plus a buffer distance to avoid collisions. Given a circular embedding with perimeter $2\pi r_d$, the maximum number of agents that could fit in a trajectory would roughly be $n_{max}\approx2\pi r_d/r_a$.

The agent dynamics follow the double integrator model~\eqref{eq:dyn}, with a white Gaussian noise of $\sigma=0.03$ added to position measurements to test robustness.

The gains used were $k_x = 6$, and $k_v = 6.5\sqrt{2}$ for the position controller~\eqref{eq:controller}, and $k_{\phi} = 0.02$ for the phase controller~\eqref{eq:phase-control}. The time step for incremental rotation updates~\eqref{eq:rot_z} was $dt=0.1$ s.

The radius of the circular embedding trajectory was $r_d=10$~m, the angular velocity was $\omega_{z,d} = 1.5$~rad/s, and the height was $h=10$~m. Note that assuming that $r_a=0.2$~m, the maximum number of agents is $n_{max}\approx 314 >50$. The parametric equations governing the embedding deformation in $\mathfrak{so}(3)$, defined in \eqref{eq:skew-x-y}, were: %according to the following:
\begin{equation}
\begin{aligned}
    \omega_{x,i}(\phi_i) &= s(\cos{(\phi_i)}\sin{(\phi_i)}-\sin^3{(\phi_i)}), \\
    \omega_{y,i}(\phi_i) &= s\,\cos^2{(\phi_i)}\sin{(-\phi_i)},
\end{aligned}
\label{eq:angular-speeds}
\end{equation}
\noindent where $s=0.4$.

\begin{figure}
% \vspace{10pt}
    \centering
    \includegraphics[width=0.8\linewidth]{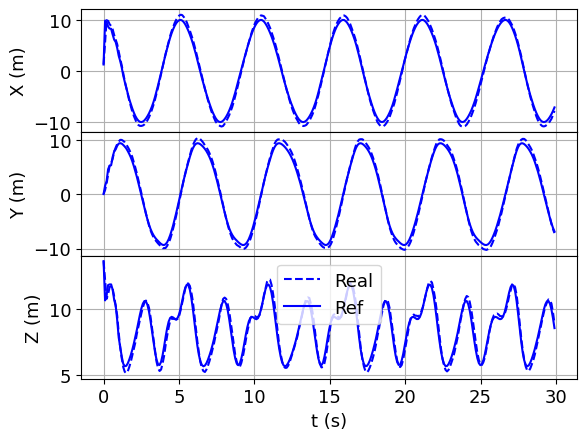}
    \caption{Desired vs. simulated positions of Agent 1, in $X$,$Y$ and $Z$ axes, as a function of time.}
    \label{fig:xyz}
\end{figure}

\begin{figure}
    \centering
    \includegraphics[width=0.8\linewidth]{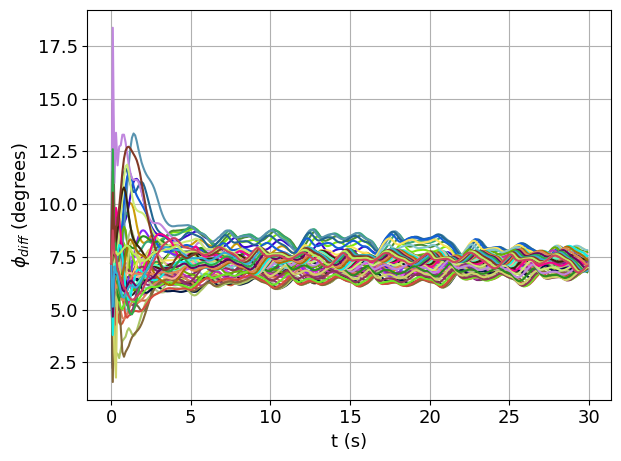}
    \caption{Phase difference between 50 agents in simulation.}
    \label{fig:phase-diff}
\end{figure}

\begin{figure}
% \vspace{10pt}
    \centering
    \includegraphics[width=0.8\linewidth]{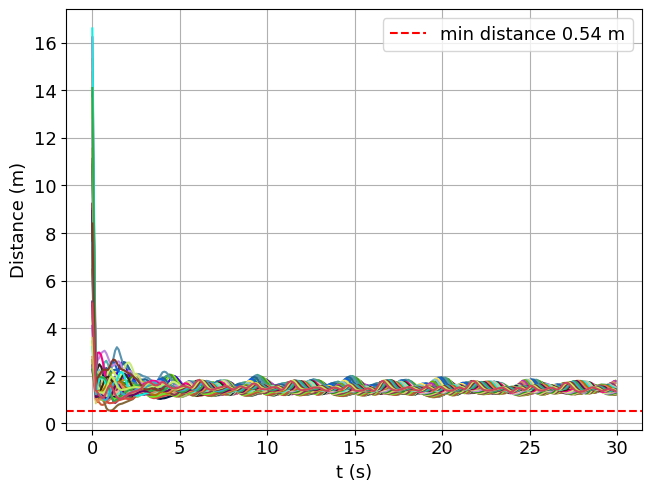}
    \caption{Cartesian distance between 50 agents in simulation.}
    \label{fig:distances}
\end{figure}

 The agents were initialized with random positions up to $10$~m away from the desired trajectory.
\begin{figure}
% \vspace{10pt}
    \centering
    \includegraphics[width=0.7\linewidth]{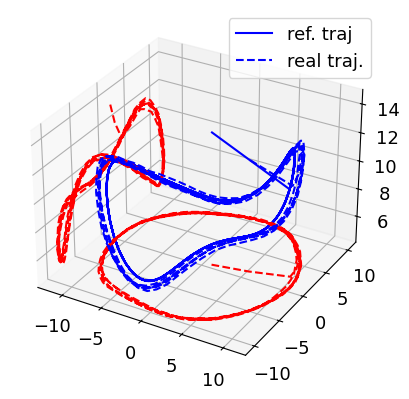}
    \caption{3D reference and simulated trajectories of Agent 1 following \eqref{eq:angular-speeds}, with $XY$ and $YZ$ projections (red dotted lines).}
    \label{fig:3D}
\end{figure}
Fig. \ref{fig:3D} illustrates the resulting 3D trajectories, showing both the reference and the executed paths. As seen in Fig.~\ref{fig:xyz}, which plots the components $X$, $Y$, and $Z$ over time, agent~1 quickly converged to the desired trajectory. This figure also shows that the deformation defined in \eqref{eq:angular-speeds} yielded $z$ coordinates ranging from approximately $5$~m to $11$~m. These results confirm the controller's ability to handle large initial deviations while maintaining robustness to noise.
Fig.~\ref{fig:phase-diff} shows that the phase controller~\eqref{eq:phase-control} maintained a target phase separation of $7.2^\circ$ between agents, as expected. The system converged to this value within approximately $5$~s. Oscillations of up to $0.8^\circ$ amplitude were observed, representing just $11\%$ of the target separation. These results confirm the controller's effectiveness in achieving precise phase regulation with numerous agents in the presence of noise. 
Considering an agent radius plus buffer distance of $r_a=0.24$~m (consistent with the platform used in the real experiments), the minimum distance for collision avoidance was ensured by the controller, as shown in Fig.~\ref{fig:distances}. Inter-agent distances ranged from $16.5$~m to $0.54$~m. The steady-state behavior was reached within the initial $5$~s

\subsection{Simulation Inserting One Agent}
In this simulation, the swarm was initiated with three agents. A fourth agent was added after $15$~s to analyze how long the phase controller would take to restore equilibrium. The trajectory parameters were $k_x=6$, $k_v=6.5\sqrt2$, $k_\phi=3$, $\omega_{z,d}=1$~rad/s, $r_d=3$~m, and $h=1.5$~m. The embedding deformation followed \eqref{eq:skew-x-y}.

\begin{figure}
    \centering
    \includegraphics[width=0.75\linewidth]{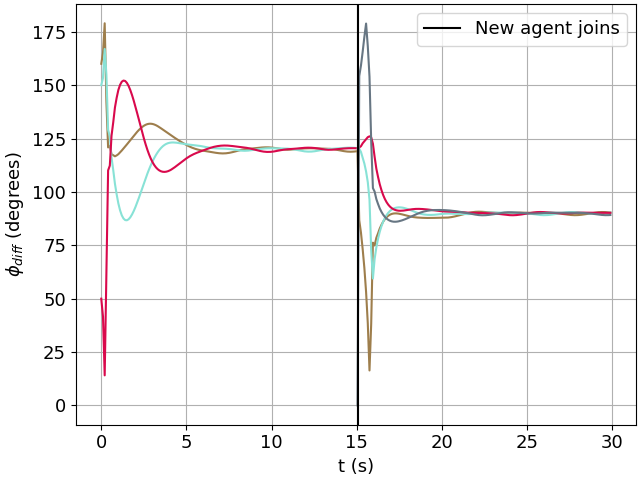}
    \caption{Angular separations between agents in the simulation initiated with three agents and one agent added at $15$~s}
    \label{fig:phase-diff-insert}
\end{figure}

\begin{figure}
    \centering
    \includegraphics[width=0.75\linewidth]{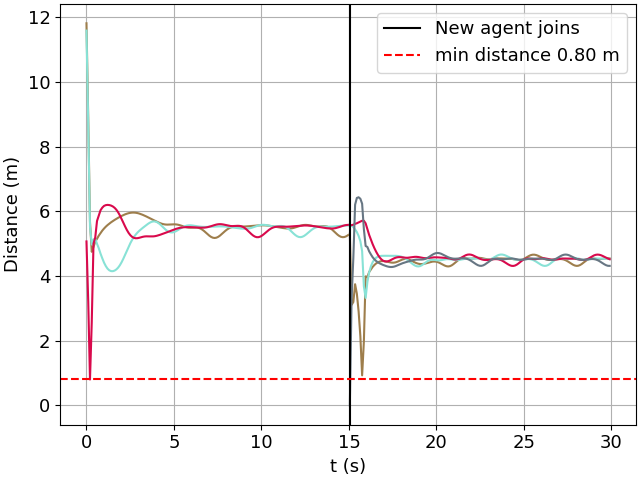}
    \caption{Distances between agents in the simulation initiated with three agents and one agent added at $15$~s}
    \label{fig:dist-insert}
\end{figure}

Fig.~\ref{fig:phase-diff-insert} shows the angular separations between agents. Initially, with three agents, the phase controllers took $5$~s to converge to the uniform phase separation ($120^\circ)$. After the insertion of the fourth agent, the controllers self-adjusted within $2.5$~s, thereby accommodating all agents with a new desired angular of $90^\circ$.
Fig.~\ref{fig:dist-insert} shows a minimum separation between agents of $0.8$~m. Considering $r_a=0.24$~m, as defined in Section~\ref{sec:sim-results}, this means collision between agents was avoided throughout the simulation.
These results demonstrate our technique automatically re-partitions the phase when an agent is added, with no centralized coordination or reparameterization required.

\subsection{Physical Experiment}\label{sec:real-results}
For the physical experiment that follows, the parametric equations for deformation at \eqref{eq:angular-speeds} and the parameter $s$ were kept identical to those used in the previous simulations. This design choice was intended to demonstrate that the proposed algorithm consistently generates the desired positions, regardless of the agent's underlying dynamics. However, the desired angular velocity was reduced to $\omega_{z,d} = 0.8$ rad/s -- lower than in the simulation -- for safety reasons. When operating multiple Crazyflie UAVs simultaneously, the radio communication channel can become congested, occasionally causing one or more drones to lose communication and control. Operating at lower speeds provides a safety margin, allowing time to trigger emergency stop procedures and prevent potential collisions. Although our technique is decentralized, all information exchanged between Crazyflies and the ground station is done via the same radio channel, due to Bitcraze's design choice.

The experiment was conducted with five Crazyflies units. Initially, all agents were placed at $y>0$, with randomly-distributed $x$ and $z$ values between $-2$~m and $2$~m. The phase controller gain was set to $k_{\phi}=8$, the desired radius was $r_d=1$~m, the height of the circular embedding was $h=0.9$~m, and the time step remained $dt=0.1$~s, as in Section~\ref{sec:sim-results}. The UAVs operated under the manufacturer-provided low-level PID controller. 

\begin{figure}
    \centering
    \includegraphics[width=0.65\linewidth]{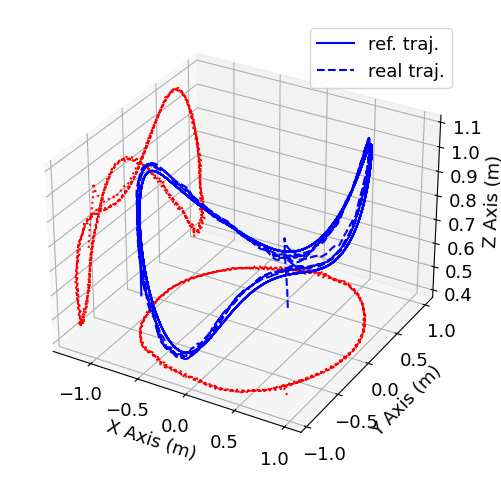}
        \caption{Desired vs. actual 3D trajectories of Quadcopter 1 and $XY$ and $YZ$ projections (red dotted lines).}
    \label{fig:real-3D}
\end{figure}

\begin{figure}
% \vspace{5pt}
    \centering
    \includegraphics[width=0.75\linewidth]{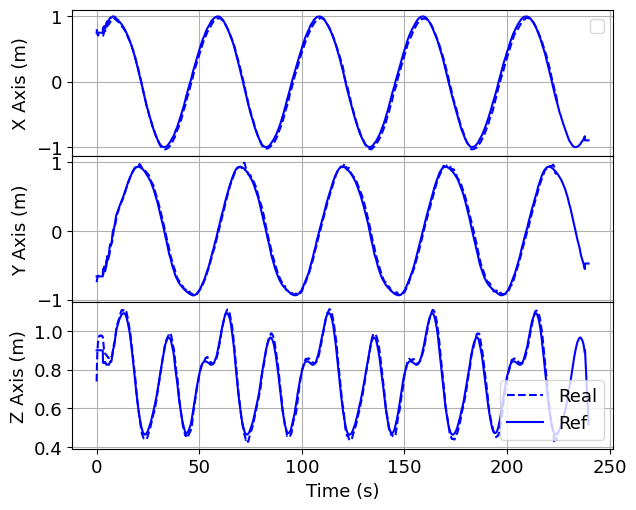}
    \caption{Desired vs. actual positions of Quadcopter 1 in $X$, $Y$, and $Z$ axes, as a function of time.}
    \label{fig:real-xyz}
\end{figure}

\begin{figure}
% \vspace{10pt}
    \centering
    \includegraphics[width=0.8\linewidth]{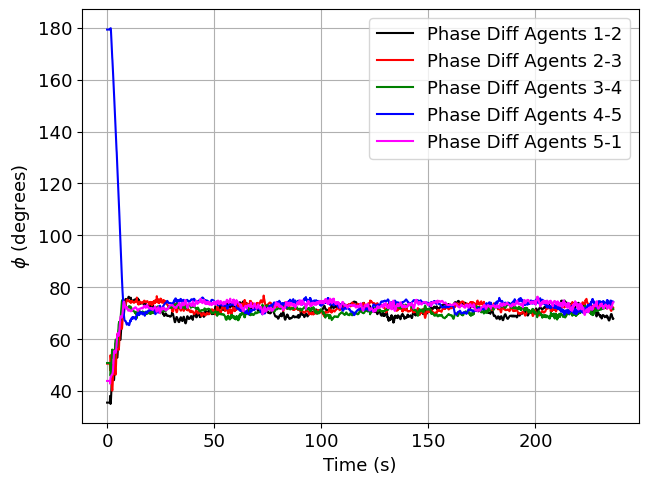}
    \caption{Phase separation between five agents, in the real experiments.}
    \label{fig:real-phase-diff}
\end{figure}

\begin{figure}
% \vspace{10pt}
    \centering
    \includegraphics[width=0.8\linewidth]{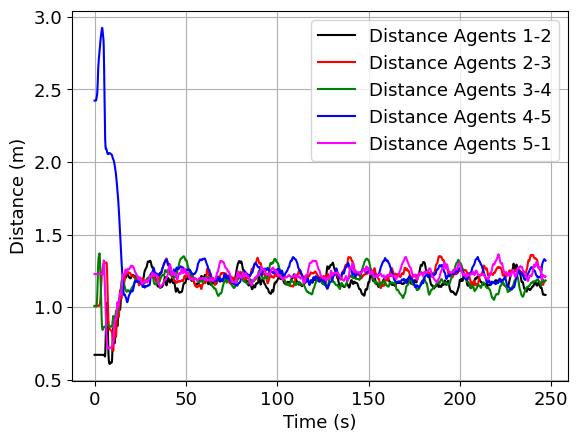}
    \caption{Cartesian distance between five agents, in the real experiments.}
    \label{fig:real-dist}
\end{figure}

Fig.~\ref{fig:real-3D} shows the 3D trajectory of Quadcopter 1 in the same view as the simulation result in Fig.~\ref{fig:3D}. The strong similarity between the two confirms that the proposed pipeline yields consistent and reliable behavior in both simulation and real-world settings, validating its applicability to physical UAV systems.%, as well as in the simulation.       
Fig.~\ref{fig:real-xyz} displays the desired and actual trajectories for one quadcopter along the $X$, $Y$, and $Z$ axes over time. A comparison with the simulation results in Fig.~\ref{fig:xyz} reveals a strong correspondence between the simulated and real-world responses across all three dimensions, confirming the consistency and accuracy of the proposed pipeline in replicating the desired behavior on physical hardware. Although Fig.~\ref{fig:real-xyz} resembles Fig.~\ref{fig:xyz} in shape, its dimensions are smaller, as the desired radius was ten times smaller. Fig.~\ref{fig:real-xyz} also shows that the agent followed the reference trajectory closely.
Fig.~\ref{fig:real-phase-diff} illustrates the effectiveness of the phase separation controller, which converged to the target angular separation of $72^\circ$ in about $10$~s and maintained it consistently throughout the experiment. Although with minor fluctuations about the desired value, the overall stability of the phase separation was preserved.
Fig.~\ref{fig:real-dist} shows that the inter-agent distances achieved a minimum value of $0.6$~m, and a maximum of $2.9$~m in the transient state. After about $15$~s, it converged and remained within $1.05$ m to $1.35$ m, confirming that collisions were successfully avoided, considering $r_a=0.24$~m. 
Periodic behavior is noticed in Fig.~\ref{fig:real-phase-diff} and Fig.~\ref{fig:real-dist} due to the periodic nature of the 3D shape.

The results in this section demonstrate that the pipeline proposed in this work can be applied to real-world agents, with more complex dynamics than a double-integrator.
\section{Conclusion}\label{sec:conclusion}
This paper introduced a novel method for generating closed 3D trajectories by deforming circular embeddings using the Lie group $\text{SO}(3)$. By leveraging the relationship between $\text{SO}(3)$ and its Lie algebra $\mathfrak{so}(3)$, we defined parametric equations directly in the algebra and converted them into rotation matrices, allowing for a richer class of closed curves than previously possible. 

Our approach requires only position inputs, eliminating the need for velocity inputs, and reducing computational overhead. This design is well-suited for resource-constrained platforms such as Crazyflies quadcopters.

We also proposed a novel phase controller that maintains uniform angular separation (\textit{i.e}, $360^\circ/n$, where $n$ is the number of agents) among agents. A formal Lyapunov stability analysis confirmed the robustness of this controller in maintaining collision-free formations.

The method was validated through simulations involving 50 agents with high angular velocity, demonstrating scalability and fast convergence even under dispersed initial conditions. Physical experiments with five quadcopters further confirmed the algorithm's effectiveness under unmodeled dynamics and with a different low-level controller architecture. Those results proved our technique to be agnostic to the agent and the low-level controller implied. % than used in simulation.

As future work, we aim to implement fully decentralized agent recognition, enabling each UAV to autonomously identify its leading and lagging neighbors without relying on an external coordinator. Furthermore, the method could be extended by adding orientation control for vehicles such as fixed-wing aircraft. 

Additionally, our work assumed perfect position measurements. In future work, we will relax this assumption and encode the measurement noise and uncertainties in the Lie algebra, as we mentioned in Section~\ref{sec:so(3)}, obtaining a more robust trajectory generation method.
\bibliographystyle{IEEEtran}
\bibliography{references}


%\vfill


\end{document}